\documentclass[letterpaper]{article} 
\usepackage{aaai2027}
\nocopyright

\usepackage[hyphens]{url}  
\usepackage{graphicx} 
\usepackage{natbib}  
\usepackage{caption} 
\usepackage{algorithm}
\usepackage{algorithmic}
\usepackage{booktabs}
\usepackage{multirow}
\usepackage{xcolor}
\definecolor{ablred}{RGB}{205,45,75}
\usepackage{newfloat}
\usepackage{listings}
\DeclareCaptionStyle{ruled}{labelfont=normalfont,labelsep=colon,strut=off} 
\floatstyle{ruled}
\newfloat{listing}{tb}{lst}{}
\floatname{listing}{Listing}

\usepackage{booktabs}

\title{MemArbiter: Decision-Time Memory Arbitration for Long-Horizon LLM Agents}
\author{
Jiajun Dong\textsuperscript{*},
Yutao Hu\textsuperscript{*},
Fengrui Fan\textsuperscript{*}, 
Shihan Dou\textsuperscript{\textdagger}, 
Yueming Wu\textsuperscript{*},
Deqing Zou\textsuperscript{*}
}

\affiliations{
\textsuperscript{*}Huazhong University of Science and Technology, China\\
\textsuperscript{\textdagger}Fudan University, China
}

\begin{document}

\maketitle

\begin{abstract}
Large language model (LLM) agents must retain and use cross-step information to act coherently in long-horizon tasks. Existing methods improve memory accessibility, yet action-relevant information may still fail to guide the current decision because it is poorly formed, organized, prioritized, or presented. We call this post-access failure the \emph{Memory-Action Gap}.
We propose {\emph{MemArbiter}}, a function-aware memory arbitration framework that addresses the memory-management-induced component of this gap. \emph{MemArbiter} decomposes interaction histories into atomic items, organizes them into five functional Memory Banks, and combines bank-level demand, item-level relevance, focal--ambient representations, and a temporal presentation gate to dynamically control memory salience.
We evaluate \emph{MemArbiter} on ALFWorld against Flat Retrieval and Flat Recency under unified per-step memory budgets. With an open-weight action-generation model, \emph{MemArbiter} achieves success rates of 82.8\% and 92.5\% under 500- and 750-token budgets, outperforming the strongest baseline by 20.9 and 25.4 percentage points, respectively. It also improves post-failure recovery and reduces failed-action repetition and state--action recurrence. These results show that function-aware memory arbitration enables accessible information to guide actions more effectively.
\end{abstract}
\section{Introduction}

Large language model (LLM) agents are increasingly applied to
long-horizon tasks, including web navigation, embodied interaction,
software engineering, and multi-turn tool use~\cite{zhou2024webarena,
yang2024sweagent, shridhar2020alfworld, farn2023tooltalk}.
As observations, actions, and execution outcomes accumulate, agents
must continually identify historical information relevant to the
current decision. The challenge lies not only in retaining such
information, but also in ensuring that it guides action selection at
the right time and in an appropriate form.

Existing studies primarily improve memory accessibility by extending or
managing usable context~\cite{peng2023yarn, packer2023memgpt},
retrieving and organizing historical
information~\cite{park2023generative}, compressing long
inputs~\cite{jiang2024longllmlingua}, or maintaining episodic
experience~\cite{shinn2023reflexion}. These approaches improve agents'
ability to preserve and revisit interaction history. However,
accessibility does not guarantee decision influence: even information
relevant to the correct action may be poorly prioritized or presented
after being stored or retrieved, and thus exert insufficient influence
on action selection~\cite{liu2024lost}.

We refer to this phenomenon as the \emph{Memory-Action Gap}:
action-relevant information is retained and accessible to the
memory-management pipeline, yet fails to guide action selection. Here,
accessibility means that the memory manager can process the information
when constructing the current decision context; it need not already be
visible in the action-generation prompt. Unlike forgetting, in which
information is not preserved, or retrieval failure, in which retained
information cannot be located or accessed, the \emph{Memory-Action
Gap} arises after information becomes available to the memory-management
process. The gap may stem from either memory management or limitations
in the underlying model's reasoning. This work focuses on the former,
specifically on how information is formed, organized, prioritized, and
presented.

A key source of the \emph{Memory-Action Gap} is the mismatch between
the heterogeneous decision functions of memories and function-agnostic
memory management~\cite{sumers2024cognitive}. In long-horizon tasks,
historical information can influence action selection in qualitatively
different ways. Goals specify the desired outcome, task states describe
current progress, constraints restrict permissible behavior, episodic
traces record previous attempts and outcomes, and reference facts
support action execution. These memories differ in persistence,
activation timing, and required salience, making homogeneous management
ill suited to their distinct decision requirements.

Supporting such differentiated management requires memory units that
can be independently updated and presented. A raw observation, however,
may simultaneously contain state changes, action outcomes, constraints,
and reference facts. When stored as an indivisible text block, these
components cannot be independently updated, prioritized, or scheduled.
We therefore formulate memory management for long-horizon agents as an
end-to-end arbitration problem spanning memory formation, functional
organization, and prompt presentation. Beyond preserving historical
information, the system must construct atomic memory units, identify
their primary functions, and dynamically adjust their priorities and
presentation states as the task evolves.

To address this problem, we propose \emph{MemArbiter}, a function-aware
dynamic memory arbitration framework for long-horizon LLM agents.
\emph{MemArbiter} decomposes interaction histories into atomic memory items and
organizes them into five functional Memory Banks: Goal, Task State,
Constraint, Episodic, and Reference. Each item can be rendered in either
a detailed focal form or a compact ambient form. At each decision step,
bank-level demand and item-level relevance signals estimate which memory
roles and items are most useful for the current decision. A temporal
presentation gate then updates each item's presentation state through
promotion, demotion, retention, or hiding. The resulting focal and
ambient memories are assembled into a structured prompt under the
available memory budget.

We implement \emph{MemArbiter} within a ReAct agent and enforce the same
per-step memory-presentation budget across all methods. On ALFWorld, we
compare \emph{MemArbiter} with Flat Retrieval and Flat Recency under two memory
budgets, conduct component ablations, and analyze its responses to
failed executions and recurrent behavior. Using the open-weight
action-generation model, \emph{MemArbiter} achieves success rates of 82.8\%
and 92.5\% under the 500- and 750-token budgets, outperforming the
strongest baseline by 20.9 and 25.4 percentage points, respectively. It
also yields higher one-step recovery rates after failed executions and
substantially reduces failed-action repetition and state-action
recurrence.

The main contributions of this work are as follows:

\noindent $\bullet$ We introduce the \emph{Memory-Action Gap}, a
post-access failure mode distinct from forgetting and retrieval failure,
in which action-relevant information is accessible to the
memory-management pipeline but fails to guide action selection.

\noindent $\bullet$ We formulate the memory-management-induced component
of this gap as end-to-end memory arbitration and propose MemArbiter,
which dynamically controls the timing, granularity, and prompt salience
of functionally organized memories under limited budgets.

\noindent $\bullet$ We evaluate \emph{MemArbiter} on ALFWorld using both
open-weight and proprietary action-generation models, showing improved
task success and failure recovery, together with fewer recurrent
actions, under unified memory budgets.
\section{Preliminaries and Problem Formulation}
\subsection{Stepwise Agent Setting}
We consider a language-model agent that interacts with an environment step by step following the ReAct~\cite{yao2022react} paradigm. 
Given a task goal \(g\), at step \(t\), the agent receives the current observation \(o_t\) and generates an action \(a_t\) based on the cross-step information maintained by a memory module \(M_t\). After executing \(a_t\), the environment returns \((o_{t+1}, r_t)\), and the memory module updates \(M_t\) with newly observed states, interaction outcomes, and other relevant information.
Before each action generation, the memory module selects and organizes information from \(M_t\) into a prompt-visible memory context \(P_t\). 
The action generation process is formulated as \(a_t \sim \pi_\theta(I, g, o_t, P_t)\), where \(I\) denotes fixed system instructions, action specifications, and output requirements, and \(\pi_\theta\) denotes the underlying language model. This work focuses on how accumulated information in \(M_t\) is organized into \(P_t\), while keeping the base model, environment, and ReAct interaction loop unchanged.

\begin{figure}[t!]
    \centering
    \includegraphics[width=0.46\textwidth]{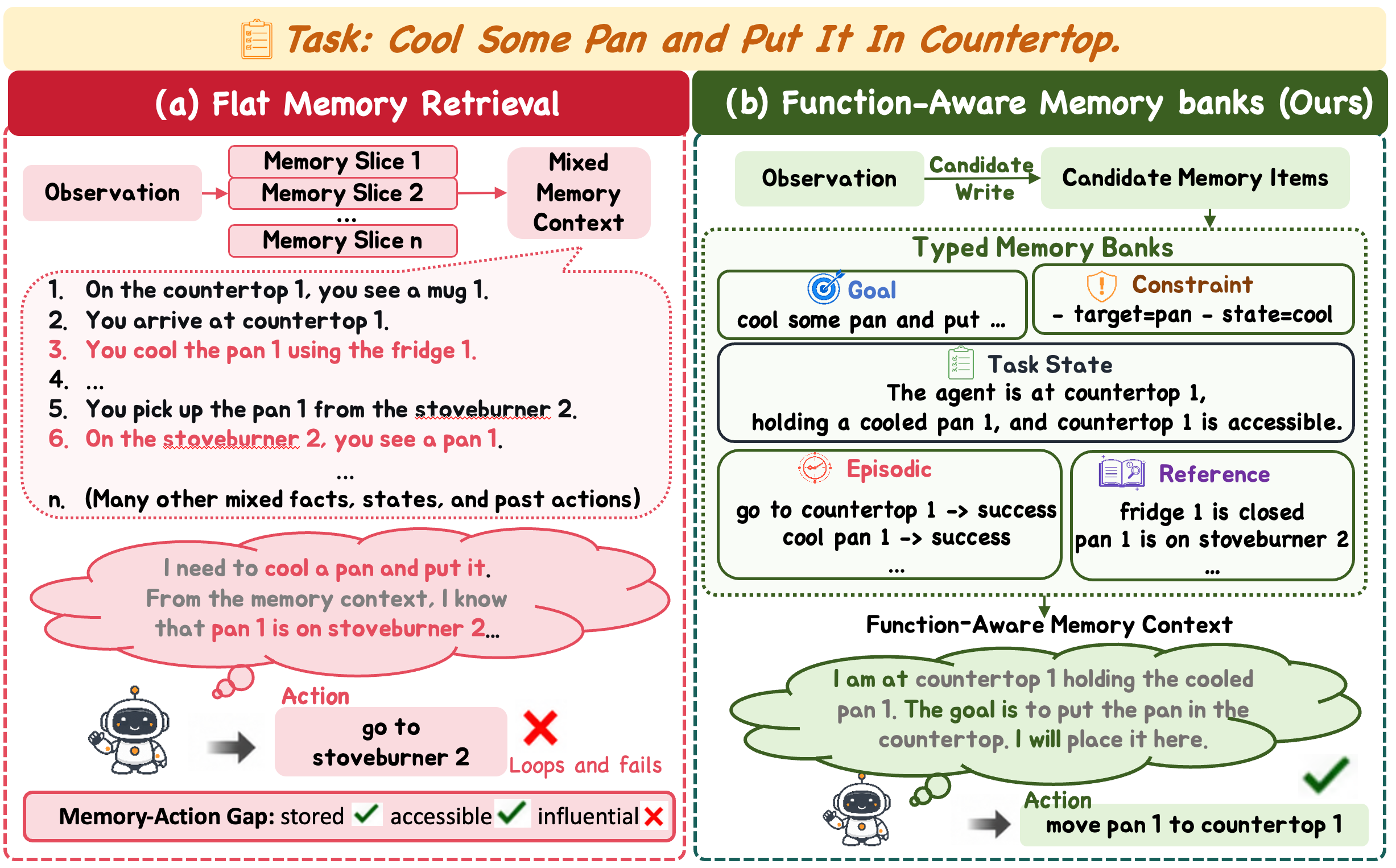}
	\caption{A case illustrating the Memory-Action Gap caused by flat memory organization. Despite relevant memories being available, flat organization mixes heterogeneous memories and weakens their action influence. Function-aware memory organization separates memories by decision functions to construct structured contexts for long-horizon agents.}
	\label{fig:pre}
\end{figure}

\subsection{Memory Items and Typed Roles}
In long-horizon tasks, an agent often cannot make appropriate decisions based solely on the current environmental observation. 
Given the current situation, the agent needs not only to identify the task objective and understand its current state, but also to reason based on accumulated experiences and environmental facts discovered during exploration. 
This cross-step information is not obtained from a single observation but accumulates throughout task execution.

\begin{table*}[t]
\centering
\caption{Functional roles and temporal characteristics of memory items.}
\label{tab:memory_roles}
\resizebox{\linewidth}{!}{
\begin{tabular}{llll}
\bottomrule
\textbf{Role} & 
\textbf{Decision Function} & 
\textbf{Representative Information} & 
\textbf{Temporal Characteristic} \\
\hline
Goal &
Guide actions toward the desired objective &
Task goals, completion criteria &
Persistent across steps \\

Task State &
Describe the current decision condition &
Location, possessed objects, object states &
Frequently updated \\

Constraint &
Restrict invalid actions and outcomes &
Attribute requirements, location constraints, quantity limits &
Stable across relevant steps \\

Episodic &
Record previous attempts and outcomes &
Executed actions, success/failure results &
Continuously accumulated \\

Reference &
Provide reusable environmental knowledge &
Object locations, identifiers, URLs, parameters &
Activated when needed \\
\bottomrule
\end{tabular}
}
\end{table*}

We define a memory item as an atomic unit of information extracted from task descriptions, environmental observations, and interaction outcomes that can support future decisions. 
Different memory items may serve fundamentally different roles in decision-making. 
For example, "a pan exists on the countertop" helps the agent recall an environmental fact discovered from previous exploration, whereas "the pan needs to be cooled" serves as a constraint that prevents the agent from performing actions violating task requirements. 
However, existing memory mechanisms often represent retrieved memories as homogeneous context. 
As illustrated in Figure~\ref{fig:pre}(a), when memory items with different decision functions are mixed together, even critical memories required for the correct decision may fail to sufficiently influence action selection despite being available in the current context, leading to a Memory-Action Gap.

To mitigate the Memory-Action Gap, we organize memory items according to their primary functions in subsequent decision-making, allowing critical memories to be appropriately utilized during current decisions while reducing interference from irrelevant memories.

Specifically, as summarized in Table~\ref{tab:memory_roles}, we categorize memory items into five functional roles: \textbf{Goal}, \textbf{Task State}, \textbf{Constraint}, \textbf{Episodic}, and \textbf{Reference}. 
Each type of memory serves a distinct function in agent decision-making. 
Goal memories maintain the task objective and provide direction for current actions; 
Task State memories describe the agent's current state and help understand the current decision context; 
Constraint memories define task requirements and restrictions, preventing actions that violate task specifications or environmental rules; 
Episodic memories record previous interactions, including actions and outcomes, enabling the agent to avoid repeated failures and reuse effective experiences; 
Reference memories preserve environmental facts discovered during exploration, providing necessary information for future execution.

Beyond their decision functions, different memory roles also exhibit distinct temporal characteristics during task execution. 
Task State memories are continuously updated as the task progresses; 
Episodic memories accumulate throughout interaction; 
Constraint memories may remain effective across multiple steps; 
and Reference memories are typically reused only when corresponding information is required. 
Therefore, different types of memory items differ in their decision roles, update patterns, and usage conditions, making unified memory management insufficient for meeting the decision requirements of long-horizon tasks.

\section{\emph{MemArbiter}}
\begin{figure*}[t!]
    \centering
    \includegraphics[width=0.8\textwidth]{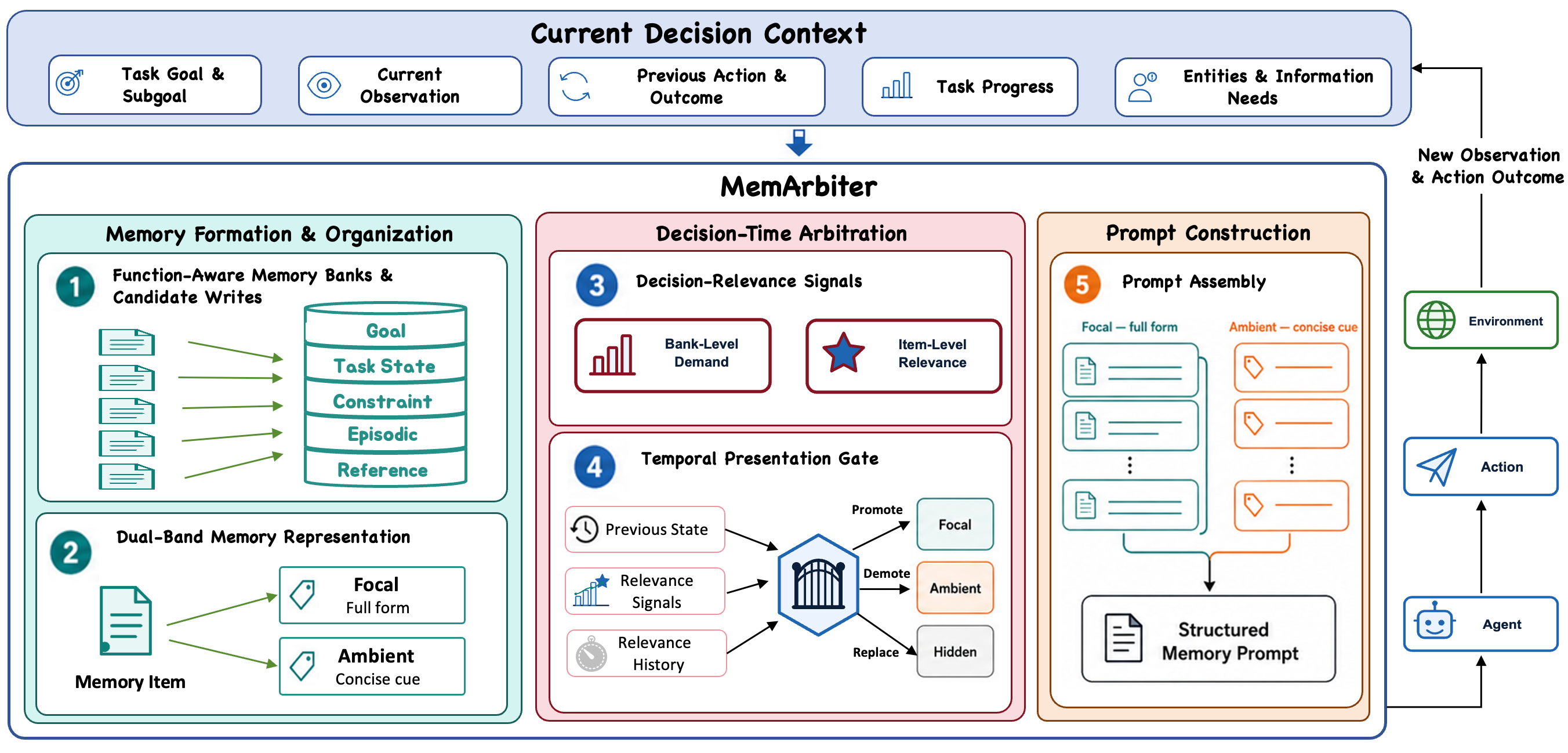}
	\caption{An overview of the \emph{MemArbiter} framework. At each decision step, \emph{MemArbiter} forms function-aware memory banks and dual-band representations, arbitrates memory presentation through decision-relevance signals and the temporal presentation gate, and assembles a structured memory prompt for action generation. New observations and action outcomes update the decision context and working memory for the next step.}
	\label{fig:overview}   
\end{figure*}

\emph{MemArbiter} is a decision-time working-memory arbitration framework for long-horizon agents. 
As shown in Figure~\ref{fig:overview}, it operates as an external memory module. 
At the beginning of step \(t\), a state-parsing step processes the task goal, current observation, previous action and its outcome, and accumulated interaction trajectory to identify the current subgoal, task progress, relevant entities, and unresolved information needs. Together, these inputs and parsed variables form the current decision context \(c_t\).

Based on \(c_t\), \emph{MemArbiter} performs five operations. It first decomposes interaction information into atomic memory candidates and updates the corresponding memory banks, while generating focal and ambient representations for each memory item. It then estimates bank-level demand and item-level relevance. Based on these signals, the Temporal Presentation Gate then combines these signals with the previous presentation state and relevance history to update each memory item's presentation state. Finally, Prompt Assembly organizes the resulting focal and ambient content into a structured memory prompt for next-action generation.

\subsection{Memory Banks and Candidate Writes}
\emph{MemArbiter} employs a Candidate Writer $\mathcal{W}$ to extract atomic information from new interaction evidence and organize it as candidate writes for the five memory banks.
At each decision step, the Candidate Writer receives the complete decision context $c_t$. The current observation $o_t$, together with the previous action and its outcome $(a_{t-1}, r_{t-1})$, provides the source content for candidate generation, while the task goal and subgoal, task progress, relevant entities, and information needs provide contextual cues for interpreting this content and assessing its value to subsequent decisions. This process is formulated as $M_t^b=\mathcal{U}_b(M_{t-1}^b,\mathcal{W}_b(c_t))$, where $\mathcal{W}_b(c_t)$ denotes the candidate writes generated for bank $b$, and $\mathcal{U}_b$ denotes its corresponding update strategy.

Each candidate write proposes either adding a new atomic memory item or revising an existing one, and includes structured attributes such as its source, related entities, and scope of applicability. The same observation or action outcome may produce candidate writes for multiple banks, while each candidate is assigned a single primary functional role.

Before updating a bank, candidate writes undergo validity checking, deduplication, and conflict resolution. Task State replaces the previous snapshot with the latest valid state. Constraint merges duplicate restrictions and revises them when their applicability or validity changes. Episodic appends temporally ordered action-outcome events that may support subsequent decisions. Reference consolidates duplicate facts and updates conflicting information based on new evidence. Depending on the update strategy, a candidate write may be rejected, merged with an existing memory, used to overwrite an outdated memory, or inserted as a new item.

\subsection{Dual-Band Memory Representation}
\emph{MemArbiter} separates memory content from its prompt-level presentation and provides two representations for each memory item $m_i$: focal and ambient. 
The focal representation retains the full content of the memory item so that it can directly support action selection, whereas the ambient representation uses a deterministic bank-specific template to produce a concise cue retaining key entities, relations, or state information. Bank and band capture two orthogonal dimensions: the former specifies a memory's decision function, while the latter determines its current presentation form. A memory item in any bank can therefore switch between the two bands without being re-extracted or reclassified. 
These representations provide alternative presentation granularities for subsequent memory arbitration, allowing the same memory to influence action generation with different levels of salience as decision needs evolve.

\subsection{Decision-Relevance Signals}
The decision value of different memory roles and individual items changes as the task progresses. \emph{MemArbiter} therefore constructs two complementary levels of signals. Bank-level demand $d_{b,t}$ estimates the overall need for functional memory role $b$ at the current step and governs whether its memories participate in the current presentation and become eligible for focal promotion. Item-level relevance $\eta_{i,t}$ evaluates the contribution of memory item $m_i$ to the current decision and guides item-level state transitions and prompt ordering.

Bank-level demand is computed from the current decision context and the functional characteristics of each bank. Goal reflects the completion status of the current subgoal and overall task progress. Task State measures the recency of the latest state snapshot. Constraint determines whether active constraints apply to the current subgoal or previous action outcome. Episodic combines whether the previous action produced an abnormal outcome with the relevance of past events to the current subgoal. Reference focuses on unresolved information needs and the relation between stored facts and the current subgoal.

Item-level relevance is computed by relating each memory item to the intermediate decision variables maintained for the current step. Goal items are evaluated by their alignment with the current subgoal, while Constraint items depend on whether their scope matches the current subgoal or previous action outcome. Task State items are evaluated by recency, Episodic items by subgoal similarity and their utility for error recovery, and Reference items by the compatibility between the information they support and the currently unresolved information needs.

\subsection{Temporal Presentation Gate}
The Temporal Presentation Gate dynamically regulates memory presentation states across consecutive decision steps to preserve the salience of critical information while adapting to changing decision needs. If the presentation form at each step were determined independently using only current relevance, transient fluctuations could prematurely remove valuable information from salient positions or cause frequent switching between bands. The Gate therefore treats presentation control as a stateful assignment problem and updates each memory item according to $z_{i,t}=\mathcal{G}{b(i)}(z{i,t-1},d_{b(i),t},\eta_{i,t},h_{i,t-1})$, where $b(i)$ denotes the bank containing $m_i$ and $h_{i,t-1}$ denotes its relevance history before step $t$. The values $F$, $A$, and $H$ of $z_{i,t}$ denote focal, ambient, and hidden states, respectively; $H$ is an exclusion state rather than a third memory representation.

The three inputs play distinct roles in state assignment. Previous state preserves presentation continuity across decision steps. Relevance signals consist of bank-level demand and item-level relevance, which respectively capture the overall need for a memory role and the decision relevance of a specific item. Relevance history distinguishes persistent trends from transient fluctuations. Based on these inputs, memories that strongly support the current decision are assigned to or retained in focal, while memories that remain useful but are less immediately relevant are presented as ambient. Memories with insufficient current relevance become hidden and are omitted from the current prompt. 
A focal memory whose relevance declines persistently is likewise demoted.

The Gate further applies role-specific transition rules. 
Goal and Constraint remain protected from ordinary temporal decay and replacement until they are completed, superseded, or invalidated. 
Task State corresponds to the latest valid state snapshot, while Episodic and Reference memories dynamically adjust their presentation states as the task progresses. 
The Gate outputs a state-annotated memory set for subsequent Prompt Assembly.

\subsection{Prompt Assembly}
Prompt Assembly is the final stage of \emph{MemArbiter}, converting the Gate output into a structured memory prompt $P_t$. Memory items are grouped into bank-specific sections to preserve their functional roles. Within each bank, focal items are presented first in full form, while ambient items use concise representations; items within the same band are ordered by item-level relevance, and hidden items are excluded.
This organization preserves both the functional structure of memory and the presentation states assigned by the Gate. The resulting $P_t$ is combined with the current observation to generate the agent's next action.
\section{Experiments}
\subsection{Experimental Setup}
\paragraph{Task Setup.}
We evaluate \emph{MemArbiter} on ALFWorld~\cite{shridhar2020alfworld}, a text-based interactive benchmark that maps embodied household tasks into multi-step language-based environments.
We use all 134 unseen tasks from the \texttt{eval\_out\_of\_distribution} split. 
Tasks require navigation and object manipulation through textual actions; at each step, the agent receives an observation and valid action templates. 
We select ALFWorld because its multi-step household tasks require agents to retain and act on information across extended interaction trajectories, making it well suited to evaluating decision-time memory management.
Episodes end upon evaluator-confirmed completion or after 50 interaction steps.

\paragraph{Baselines.}
We compare \emph{MemArbiter} with two type-agnostic baselines under the same memory-presentation budget. \emph{Flat Recency} stores observations and actions as an unstructured history and retains the most recent records within the limit. \emph{Flat Retrieval} uses BM25 \cite{robertson2009probabilistic} to rank observation lines, with a query formed by concatenating the task goal, previous action, and the latest observation, and selects the highest-ranked lines within the same limit. 
To evaluate the contribution of each component, we further evaluate \emph{w/o Functional Banks}, \emph{w/o Dynamic Signals}, and \emph{w/o Temporal Gate} under the 500-token setting.

\paragraph{Metrics.}
We report cumulative success rate, denoted as $\mathrm{SR}@k$, which is the proportion of the 134 tasks completed within $k$ interaction steps. 
$\mathrm{SR}@50$ is the primary metric, while intermediate thresholds compare completion speed over the same task set. For mechanism diagnostics, we analyze post-failure recovery, recurrent behavior, and post-failure changes in the prompt share of Episodic memory.

\paragraph{Implementation Details.}
We use Qwen3.6-27B-FP8~\footnote{\url{https://huggingface.co/Qwen/Qwen3.6-27B-FP8}} for action generation and, within \emph{MemArbiter}, for candidate writing and current-state parsing. In the cross-model experiment, GPT-5.4~\footnote{We utilized the model via OpenAI API service.} replaces it only for action generation.
Qwen3.6 is served through vLLM~\cite{kwon2023efficient} with a maximum context length of 4,096 tokens on a single NVIDIA A100-SXM4-80GB GPU.
We evaluate the three main methods under memory-presentation budgets of 500 and 750 tokens, while all ablation variants use the 500-token setting. 
These limits apply only to the memory block inserted into the action prompt and are measured uniformly using a shared budget-accounting tokenizer \texttt{o200k\_base} encoding provided by \texttt{tiktoken}.
For the Qwen3.6 experiments, each task is run once with temperature 0 and random seed 42.
For BM25, we use \(k_1=1.2\) and \(b=0.75\), with lowercase conversion and whitespace tokenization only.

\subsection{RQ1: Overall Performance}

\begin{table*}[t]
\centering
\caption{Success rate (\%) by task type on the 134-task ALFWorld
unseen split. Numbers in parentheses indicate category sizes.
The best result under each memory budget is shown in bold.}
\label{tab:alfworld_main}
\renewcommand{\arraystretch}{0.85}
\small

\begin{tabular*}{0.7\textwidth}{
@{\extracolsep{\fill}}
clccccccc
@{}
}
\toprule
\textbf{Budget}
& \textbf{Method}
& \textbf{Pick}
& \textbf{Look}
& \textbf{Clean}
& \textbf{Heat}
& \textbf{Cool}
& \textbf{Pick Two}
& \textbf{Overall} \\
&
& (24) & (18) & (31) & (23) & (21) & (17) & (134) \\
\midrule

\multirow{3}{*}{500}
& Flat Recency
& 66.67 & 83.33 & 70.97 & 21.74 & 47.62 & 0.00 & 50.75 \\

& Flat Retrieval
& 75.00 & 66.67 & \textbf{83.87}
& 52.17 & 71.43 & 0.00 & 61.94 \\

& \emph{MemArbiter}
& \textbf{95.83}
& \textbf{100.00}
& 80.65
& \textbf{73.91}
& \textbf{85.71}
& \textbf{58.82}
& \textbf{82.84} \\

\midrule

\multirow{3}{*}{750}
& Flat Recency
& 70.83 & 83.33 & 77.42 & 21.74 & 42.86 & 0.00 & 52.24 \\

& Flat Retrieval
& 83.33 & 72.22 & 83.87 & 65.22 & 76.19 & 0.00 & 67.16 \\

& \emph{MemArbiter}
& \textbf{100.00}
& \textbf{100.00}
& \textbf{93.55}
& \textbf{86.96}
& \textbf{100.00}
& \textbf{70.59}
& \textbf{92.54} \\

\bottomrule
\end{tabular*}
\end{table*}

\begin{figure}[t!]
    \centering
    \includegraphics[width=0.47\textwidth]{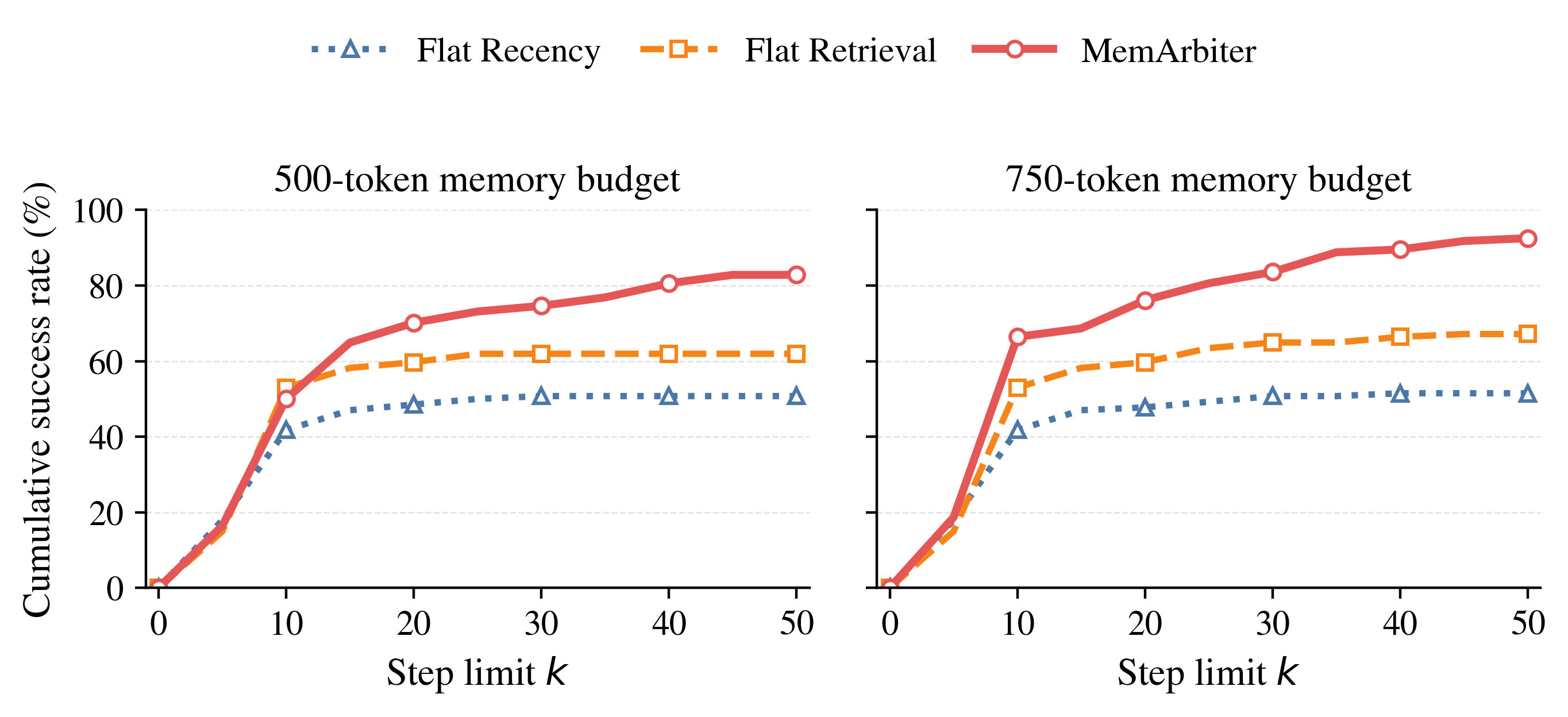}
	\caption{Success rate under different interaction-step limits on ALFWorld.
$\mathrm{SR}@k$ denotes the percentage of all 134 tasks completed within
$k$ steps. Results are reported at five-step intervals under 500- and
750-token memory budgets.}
	\label{fig:sr}
\end{figure}

Table~\ref{tab:alfworld_main} reports the success rates on the 134 unseen ALFWorld tasks. \emph{MemArbiter} consistently outperforms both flat-memory baselines under both memory budgets. With a 500-token budget, it achieves an overall success rate of 82.84\%, exceeding the strongest baseline, Flat Retrieval, by 20.90 percentage points. With 750 tokens, its success rate further increases to 92.54\%, yielding a 25.38-point improvement over Flat Retrieval. In contrast, increasing the memory budget produces only modest gains for Flat Recency and Flat Retrieval. 
This suggests that additional memory capacity is more useful when the retained information is organized and presented according to its decision function, rather than accumulated or retrieved as a flat sequence.

The task-type breakdown shows that the improvement is not confined to a single task category. Under the 500-token budget, \emph{MemArbiter} performs best on five of the six task types, with Flat Retrieval retaining a small advantage on Clean tasks. Under the 750-token budget, \emph{MemArbiter} achieves the highest success rate across all six categories. 
The largest gains occur on Pick Two tasks: both flat baselines fail all Pick Two tasks, whereas \emph{MemArbiter} reaches 58.82\% and 70.59\% success under the two budgets.
These tasks require the agent to preserve and coordinate multiple intermediate states, making them particularly sensitive to whether previously observed information remains available to subsequent decisions.

\paragraph{Success under step limits.}
Figure~\ref{fig:sr} reports $\mathrm{SR}@k$, the percentage of all tasks completed within an interaction-step limit $k$, evaluated at five-step intervals. Failed tasks remain in the denominator, so the metric jointly reflects task completion and interaction efficiency without favoring methods that succeed only on simpler tasks. \emph{MemArbiter} maintains a higher cumulative success rate than the flat baselines across most step limits, and the performance gap generally widens as more interaction steps become available. At $k=50$, the curves converge to the overall success rates reported in Table~\ref{tab:alfworld_main}. Importantly, \emph{MemArbiter} completes more tasks under the same step limit, indicating that its higher final success rate is not merely obtained by taking more actions. Instead, function-aware memory arbitration enables the agent to make productive use of longer interaction trajectories, while the flat baselines plateau substantially earlier.

\begin{table}[t]
\centering
\caption{Cross-model validation using GPT-5.4 as the action-generation model under a 500-token memory-prompt budget.}
\small
\renewcommand{\arraystretch}{0.9}
\begin{tabular}{lcc}
\toprule
Method & SR@15 $\uparrow$ & SR@50 $\uparrow$ \\
\midrule
\emph{MemArbiter}     & \textbf{70.14} & \textbf{83.58} \\
Flat Retrieval & 68.65 & 70.89 \\
Flat Recency   & 68.65 & 73.13 \\
\bottomrule
\end{tabular}
\label{tab:gpt_validation}
\end{table}


\paragraph{Cross-model validation.}
We replace only the action-generation model with GPT-5.4 under the 500-token memory budget, while retaining Qwen3.6 for candidate writing and state parsing. All tasks, prompts, and interaction limits remain unchanged.
As shown in Table~\ref{tab:gpt_validation}, \emph{MemArbiter} achieves an SR@50 of 83.58\%, outperforming Flat Retrieval and Flat Recency by 12.69 and
10.45 percentage points, respectively. 
These results suggest that the benefit of memory arbitration persists when the action-generation model is changed from an open-weight to a proprietary LLM.

\subsection{RQ2: Ablation Analysis}
Table~\ref{tab:ablation} evaluates the principal components of \emph{MemArbiter} under the 500-token memory budget. Replacing the functional Memory Banks with a single memory pool reduces $\mathrm{SR}@50$ by 15.68 percentage points, despite a relatively small 3.74-point decrease at $\mathrm{SR}@15$. This widening gap suggests that separating memories by decision function becomes increasingly important as interaction histories grow and heterogeneous information must remain distinguishable. Replacing the dynamic bank-level signals with a fixed allocation produces the largest final degradation, decreasing $\mathrm{SR}@50$ by 17.17 points. Although the static allocation remains competitive at the earlier step limit, it cannot adapt memory emphasis as the active subgoal and information requirements change.

\begin{table}[t]
\centering
\small
\renewcommand{\arraystretch}{0.9}
\caption{Ablation results on ALFWorld under the 500-token memory
budget. Parenthesized values indicate changes in percentage points
relative to full \emph{MemArbiter}.}
\begin{tabular*}{\columnwidth}{
    @{\extracolsep{\fill}}lcc@{}
}
\toprule
\textbf{Variant}
& \(\mathbf{SR@15}\uparrow\)
& \(\mathbf{SR@50}\uparrow\) \\
\midrule
\textbf{\emph{MemArbiter}}
& 64.93
& \textbf{82.84} \\
\midrule
\textit{w/o Functional Banks}
& 61.19 ($-$3.74)
& 67.16 ($-$15.68) \\
\textit{w/o Dynamic Signals}
& 62.69 ($-$2.24)
& 65.67 ($-$17.17) \\
\textit{w/o Temporal Gate}
& \textbf{67.91} ($+$2.98)
& 73.88 ($-$8.96) \\
\bottomrule
\end{tabular*}
\label{tab:ablation}
\end{table}

Removing the Temporal Gate yields a 2.98-point improvement at $\mathrm{SR}@15$ but an 8.96-point reduction at $\mathrm{SR}@50$. Independent per-step selection can favor immediately relevant items during the early trajectory, but it is less effective at preserving useful information across longer decision sequences. Overall, the results indicate that functional organization, dynamic bank-level signals, and temporal state maintenance provide complementary benefits, with their contributions becoming most apparent in longer interactions.

\subsection{RQ3: Failure Response and Recurrence}
\label{sec:mechanism_response}

For one-step recovery and failed-action repetition, we restrict evaluation to common-failure tasks, defined as tasks on which all three methods encounter at least one failed execution (29 tasks under the 500-token budget and 32 under the 750-token budget). Within each task, \emph{one-step recovery} is the fraction of failed executions whose immediately following action is successfully executed by the environment, while \emph{failed-action repetition} is the fraction whose immediately following normalized action repeats the failed action. We then macro-average these task-level rates over the common-failure task set, assigning equal weight to each task regardless of its number of failures. A successful next execution does not imply task completion or guarantee task progress. Over complete trajectories, \emph{adjacent repetition} measures identical consecutive actions, whereas \emph{state-action recurrence} checks whether the same normalized pre-action observation and action pair has appeared earlier in the episode.

As shown in Table~\ref{tab:mechanism_response}(a), at 500 tokens, \emph{MemArbiter} improves one-step recovery over Flat Retrieval by 18.2 percentage points and reduces failed-action repetition by 20.4 points.
The differences increase to 34.8 and 31.4 points at 750 tokens.
\emph{MemArbiter} also achieves the lowest adjacent repetition and state-action recurrence under both budgets, indicating fewer recurrent behaviors beyond the step immediately following a failure.

Panel (b) examines the corresponding memory response. For each failed execution, we measure the change in the proportion of the memory prompt occupied by Episodic content from the failed action to the immediately following action. This share increases by 3.3 and 1.9 percentage points at 500 and 750 tokens, respectively, while the w/o Temporal Gate variant shows little change (+0.2 points). 
These results show an association between failure-sensitive memory presentation and improved recovery and fewer recurrent actions, without implying direct causality.

\begin{table}[t]
\centering

\caption{
Panel (a) reports task-macro behavioral metrics. The two failure-response metrics use the common-failure task sets defined in the text, whereas the two recurrence metrics use all tasks.
Panel (b) shows the percentage of memory-prompt tokens occupied by
Episodic content in the prompt that produced a failed action and in the next prompt assembled after observing the failure.
}

{\scriptsize
\setlength{\tabcolsep}{2pt}
\renewcommand{\arraystretch}{0.98}

\begin{tabular*}{\columnwidth}
{@{\extracolsep{\fill}}lccc@{}}
\toprule
\multicolumn{4}{l}{
\textit{(a) Failure recovery and recurrent behavior (\%)}
}\\
\addlinespace[1pt]
\textbf{Metric}
& \textbf{\emph{MemArbiter}}
& \shortstack{\textbf{Flat}\\\textbf{Retrieval}}
& \shortstack{\textbf{Flat}\\\textbf{Recency}} \\
\midrule

\multicolumn{4}{l}{\textit{500-token budget}} \\
One-step recovery $\uparrow$
& \textbf{59.9} & 41.7 & 18.5 \\
Failed-action repetition $\downarrow$
& \textbf{26.7} & 47.1 & 80.8 \\
Adjacent repetition $\downarrow$
& \textbf{4.15} & 15.72 & 29.91 \\
State-action recurrence $\downarrow$
& \textbf{12.99} & 26.99 & 38.49 \\

\midrule

\addlinespace[1pt]
\multicolumn{4}{l}{\textit{750-token budget}} \\
One-step recovery $\uparrow$
& \textbf{65.3} & 30.5 & 16.3 \\
Failed-action repetition $\downarrow$
& \textbf{20.7} & 52.1 & 82.8 \\
Adjacent repetition $\downarrow$
& \textbf{2.33} & 15.40 & 29.96 \\
State-action recurrence $\downarrow$
& \textbf{4.33} & 24.85 & 38.64 \\
\bottomrule
\end{tabular*}

\smallskip

\begin{tabular*}{\columnwidth}
{@{\extracolsep{\fill}}lccc@{}}
\toprule
\multicolumn{4}{l}{
\textit{(b) Episodic share before and after failure (\%)}
}\\
\addlinespace[1pt]
\textbf{Setting}
& \textbf{Budget}
& \shortstack{\textbf{Failed-action}\\\textbf{prompt}}
& \shortstack{\textbf{Next}\\\textbf{prompt}} \\
\midrule
\emph{MemArbiter} & 500 & 29.4 & \textbf{32.7} \\
\emph{MemArbiter} & 750 & 31.4 & \textbf{33.3} \\
w/o Temporal Gate    & 500 & 19.9 & 20.1 \\
\bottomrule
\end{tabular*}
}
\label{tab:mechanism_response}
\end{table}

\section{Related Work}
\paragraph{Agent Memory Management.}
To enhance agent adaptability, agent memory has attracted increasing attention from a growing body of work~\cite{zhang2025survey}.
Existing agent-memory research spans persistent cross-session systems~\cite{zhong2024memorybank, xu2026A-mem, tan2025prospect, kang2025memory_os_of} and in-trial context compression or summarization~\cite{jiang2024longllmlingua, wu2025resum, zhou2025mem1, ai2026cognitive}. 
Experience-based agents retain reusable lessons, skills, programs, or workflows from prior interactions~\cite{zhao2024expel, wang2023voyager, sarch2024vlm, wang2024agent}.
Among working-memory methods, HiAgent~\cite{hu2025hiagent} organizes histories by subgoal, ARC~\cite{yao2026arc} maintains a revisable execution context, while MemAct~\cite{zhang2026memory_as_action} and AgeMem~\cite{yu2026agentic} learn generic memory-editing operations. 
These methods demonstrate the importance of actively managing in-trial information, but primarily organize working memory through subgoal progress, unified summary states, or generic editing operations.

\paragraph{Memory Utilization for Action Selection.}
Liu~\cite{liu2024lost} shows that relevant information already present in an input may be used unreliably depending on its position. 
At the agent level, Xiong~\cite{xiong2026how_memory} finds that retrieved experiences may propagate errors or provide misleading guidance.
Mem2ActBench~\cite{shen2026mem2actbench} further evaluates whether agents can ground long-term memories in executable tool calls, revealing bottlenecks in retrieval and parameter grounding. 
These findings distinguish memory accessibility from effective action guidance. 
\emph{MemArbiter} focuses specifically on within-trajectory working memory and targets the portion of this Memory-Action Gap caused by how heterogeneous memories are formed, organized, and presented at each decision step.
\section{Limitations and Future Work}
Our study has three main limitations. \textbf{(1) Single-environment evaluation.} We evaluate MemArbiter on ALFWorld, which consists of long-horizon, text-based household tasks. The results therefore establish its effectiveness within this setting, but do not show whether the same arbitration behavior transfers to web, multimodal, or more open-ended environments. We leave evaluation in these
environments to future work.
\textbf{(2) Explicit role design.} The five Memory
Banks and their relevance signals are defined in advance, enabling transparent and auditable arbitration across distinct decision functions. Future work may investigate domain-adaptive roles and signals while preserving this functional structure. 
\textbf{(3) Associative
mechanism evidence.} Our diagnostics relate memory presentation patterns, such as increased Episodic exposure after failure, to subsequent agent behavior.
They do not establish whether an individual promotion or demotion directly causes the resulting action. Controlled interventions on memory presentation states are needed to identify such causal effects.
\section{Conclusion}

We introduced the Memory--Action Gap, in which decision-relevant information retained by an agent fails to exert sufficient influence on action selection, and proposed \emph{MemArbiter} to regulate how such information participates in decision-time prompting. Under unified memory budgets on ALFWorld, \emph{MemArbiter} consistently improves task success over flat memory-management baselines across the evaluated budgets. Component ablations and mechanism diagnostics further show higher recovery following failed executions, fewer repeated and recurrent actions, and an increased Episodic memory share immediately after failure. These findings support the importance of memory arbitration beyond storage and retrieval alone. By coordinating functional memory roles, decision relevance, and temporal presentation states, \emph{MemArbiter} provides a practical foundation for more adaptive memory use in long-horizon agents.

\bibliography{aaai2027}

\end{document}